\documentclass[journal]{IEEEtran}
\usepackage{wrapfig}
\usepackage{cite}
\usepackage{amsmath,amssymb,amsfonts}
\usepackage{multirow}
\usepackage{graphicx}
\usepackage{textcomp}
\usepackage{xcolor}
\usepackage{times}
\usepackage{soul}
\usepackage{url}
\usepackage{graphicx}
\usepackage{amsmath}
\usepackage{subfigure}
\usepackage{float}
\usepackage{bm}
\usepackage{amsmath,amssymb,amsfonts,amsthm}
\usepackage{booktabs}
\usepackage{graphicx}
\usepackage{amssymb}
\usepackage{amsmath}
\usepackage{amsthm}
\usepackage{booktabs}
\usepackage[ruled]{algorithm2e} 
\usepackage{algorithmicx}
\usepackage{algpseudocode} 
\usepackage{graphicx}
\usepackage{subfigure}
\usepackage{color,xcolor}
\ifCLASSINFOpdf
\else
\fi
\begin{document}

%
\title{What You See is What You Grasp:
	User-Friendly Grasping Guided by
	Near-eye-tracking}
%

%
%

\author{ Shaochen Wang$^*$, Wei Zhang$^*$, Zhangli Zhou$^*$, Jiaxi Cao, Ziyang  Chen, Kang Chen, Bin Li, Zhen Kan$^{}$
\thanks{* Contribute equally }
\thanks{The authors are with the School of Information Science, University of Science and Technology of China, Hefei, 230026, China.}
}

%
%

\maketitle
\thispagestyle{empty}
\pagestyle{empty}

%



\begin{abstract}
This paper presents a next-generation human-robot interface that can infer and realize the user's manipulation intention via sight only. Our system integrates near-eye-tracking with robotic manipulation to enable user-specified actions, such as grasping or pick-and-place operations. We develop a head-mounted near-eye-tracking device that tracks the user's eyeball movements in real-time to identify the user's visual attention and enable sight-guided manipulation. To improve grasping performance, we introduce a transformer-based grasp model that uses stacked attention blocks to extract hierarchical features, expanding channel volumes while squeezing feature map resolutions. Experimental validation demonstrates that our system can effectively assist users to complete manipulation tasks through their eyes, which holds great potential for an assistive robot that leverages gaze interaction to aid individuals with upper limb disabilities or the elderly in their daily lives.

\end{abstract}


%
\IEEEpeerreviewmaketitle

\section{Introduction}

Robotic systems \cite{Sol22},\cite{9810182}, powered by recent advances in artificial intelligence, have become increasingly prevalent in both industrial and  daily life.  To provide more daily assistance, robots are no longer limited to performing simple and repetitive tasks but are designed to understand human intention for better helping users. In particular, it is essential to build bridges between humans and machines to transfer human intentions to robots.

Traditional  manipulation methods \cite{9087500} often involves using a joystick to deliver user-specified actions.  Unfortunately, this can be challenging for elderly individuals, particularly those with upper limb disabilities. Recent advancements in wearable technology have demonstrated the potential in robotic assistance systems such as brain-computer interface (BCI)\cite{SunZCWTS19}. However, invasive BCI devices require microelectrodes to be surgically implanted into the human cerebral cortex, which can be risky. Furthermore, non-invasive BCI devices are susceptible to noise interference and are often expensive.
Therefore, there is a pressing need to develop a new human-robot interface that is both safe and user-friendly.

The majority of human sensory input is acquired through the eyes. In fact, over 80\% of human information is obtained through vision \cite{rosenblum2011see}, which motivates the development of eye-based robotic assistive systems to enable manipulation using sight.
Robotic assistive systems that incorporate eye-tracking technology have the potential to revolutionize various applications, including surgical diagnosis, rehabilitation, and academic research. For individuals with physical disabilities, eye-tracking offers a natural interface that can bridge the gap between humans and robots, since the user's vision is typically not affected by their disability. However, despite its potential benefits, eye-tracking has not been widely adopted in practice, largely due to the difficulty of accurately modeling the motion of the eye to capture the gaze point. Currently, many eye-tracking robotic assistive systems \cite{cio2019proof} rely on fixed cameras and are primarily built for desktop use only.

This paper introduces a next-generation human-robot interface that can infer and execute user's manipulation intention using only sight. We have developed a system that integrates near-eye-tracking and robotic manipulation to enable user-specified actions, including grasping and pick-and-place tasks. To achieve sight-guided manipulation, we have designed a head-mounted near-eye-tracking device that tracks eyeball movements in real-time, enabling us to identify the user's visual attention. Additionally, we have developed a transformer-based global grasp detection framework that enhances the robot's sensing capabilities, facilitating successful execution of user-specified manipulations. Our framework utilizes self-attention to model the long-term spatial dependencies among pixels, and a feature fusion pyramid to merge multi-scale features from each stage, thereby determining the final grasping pose. Experimental validation has demonstrated that the eye-tracking system has low gaze estimation error, and the grasping system performs well on multiple grasping datasets.

The contribution of this paper can be summarized as follows:
\begin{itemize}
	\item  	We build a sight-based robotic assistive system for user-specified manipulations. Our system includes a head-mounted device that enables real-time intention inference and a grasping subsystem that utilizes self-attention mechanisms for improved visual grasping.
	
	\item  We propose a novel human-robot interface that enables more natural and instinctive manipulation by utilizing eye-tracking only.
	
	\item Extensive experiments demonstrate the effectiveness of the developed robotic assistive system for manipulation tasks.
	
\end{itemize}

\section{Related Work}
Robotic manipulation \cite{6476697},\cite{kumra2017robotic}, \cite{9959267} is a fundamental skill that has found widespread applications in manufacturing, industry, and medical operations. Vision-based grasping techniques have been extensively investigated by researchers. Lenz et al. \cite{lenz2015deep} were the first to utilize deep learning to detect grasping rectangles. Redmon et al. \cite{redmon2015real} employed a convolutional neural network (CNN) to regress the grasping pose that the robot can execute. Additionally, Morrison et al. \cite{morrison2020learning} designed a generative grasping CNN (GG-CNN) that uses depth input to generate antipodal grasp candidates. With the recent advances in artificial intelligence, the new generation of robots is expected to understand human intention through interactions with users, rather than being limited to low-level tasks such as grasping.

Assistive robotic arms have become increasingly popular among users with upper limb impairments in their daily lives, such as grasping objects and pouring water. While joysticks are often used to control these robots, they can be difficult to use, especially for elderly or upper limb disabled users. In contrast, sight is a natural way for people with physical disabilities or mobility and speech impairments to interact with robotic systems. Eye-tracking technology, which has been used for almost a century, has advanced significantly in recent years and can now be used to manipulate robotic devices by following human gaze.

Note that the attention mechanism mimics the way how human vision works. Since sight can indicate human attention, Hollenstein et al. \cite{HollensteinZ19} enhanced the performance of the  annotation model by using human sight, and demonstrated that the semantic information embedded in the sight can be well utilized by the entity model. In computer vision tasks, Karessli et al. \cite{karessli2017gaze} improved the classification accuracy of zero-shot tasks by introducing human sight  as an auxiliary task. In vision-language tasks, Sugano et al. \cite{sugano2016seeing} assisted the image caption tasks with human sight annotation information.
In addition, eye-tracking technology has been used in augmented reality \cite{5772406}, mixed reality \cite{bruno2020mixed}, and deep learning  \cite{sugano2016seeing}. These successful applications motivate  our research of using sight to enable 
human-robot manipulation.

\section{Method}

\subsection{System Overview}
\begin{figure}[]
	\centering:
	\includegraphics[height=0.34\textwidth]{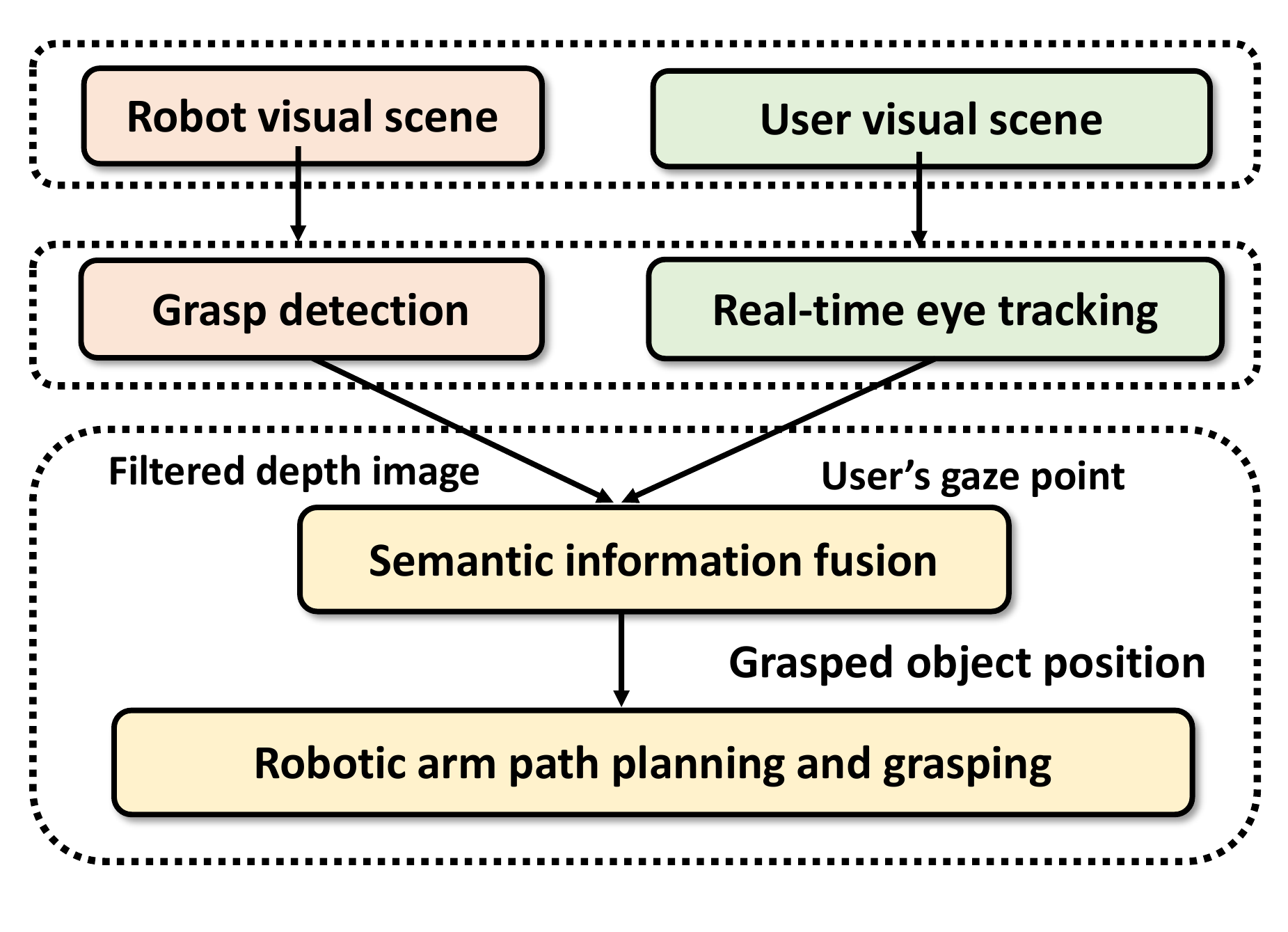} 
	\caption{The whole system pipeline. The pink section indicates the grasp detection subsystem and the green section is the eye-tracking module. The bottom yellow area is the fusion module, where objects of interest are selected for grasping based on the user's gaze. }
	\label{pipeline}
\end{figure}

This section presents a friendly human-robot interaction by incorporating
near-eye-tracking with robotic manipulation.
The developed robotic assistive system is integrated with a head-mounted eye-tracking device so that the user's intention can be conveniently captured. That is, the user can use the sight to control a robotic arm to manipulate and grasp objects. Fig. \ref{pipeline} illustrates the pipeline of how the human intentions are perceived through eye gaze and translated to actions that the robot can execute. The method involves a combination of three steps: \romannumeral1) A head-mounted eye-tracking device integrated with low-cost stereo cameras measures the user's gaze direction. The biological model of human eyeball is incorporated with computer vision to identify the pupil orientation and locate the gaze coordinates of the eyes in 3D space. \romannumeral2) In parallel, a hierarchical transformer visual model is developed to extract effective features for grasping, where the attention performs global perception. A feature pyramid inside the transformer network gathers features from each stage for multi-scale sensing in order to generate the final grasping configuration. \romannumeral3) The information from the two subsystems is fused and the gaze point with human attention is filtered against the grasping quality heatmap to obtain the grasping pose parameters for the desired object.
\subsection{ Eye-tracking}
The human eye is a highly sophisticated optical instrument that allows light entering the eye to be imaged on the retina through a series of reflections and refractions. The refractive system of eyes   consists of the cornea, vitreous humor, and lens, which forms multiple refractive surfaces at their interfaces due to the different refractive indices of each part, making the optical system extremely complex. In practice, to simplify the model, the refraction of the cornea during imaging of pupil is ignored and the cornea is modelled as a sphere with the same curvature at all points. We employ the eyeball model presented by \cite{nagamatsu2010gaze}, where the eye is modelled as two nested ellipsoids. The larger ellipsoid is the  ocular and the smaller  is the cornea. The corneal surface is modelled  as a rotating ellipsoidal plane.
The pupil is the channel for light to enter the eye and the direction of pupil directly indicates the rotation of eyeball and sight. The central pupil coordinate is the most important feature in  sight tracking. Our pupil central  localization is split into two
steps: coarse localization and refined localization.  The coarse localization mainly uses radical symmetry transformations to quickly locate points in the pupil area and eliminate invalid cases such as blink frames. Refined localization is  carried out by edge extraction using Canny operation, followed by edge filtering and ellipse fitting of edge segments to obtain the pupil's precise position.

\begin{figure}
	\centering	
	{
		\includegraphics[width=0.42\textwidth]{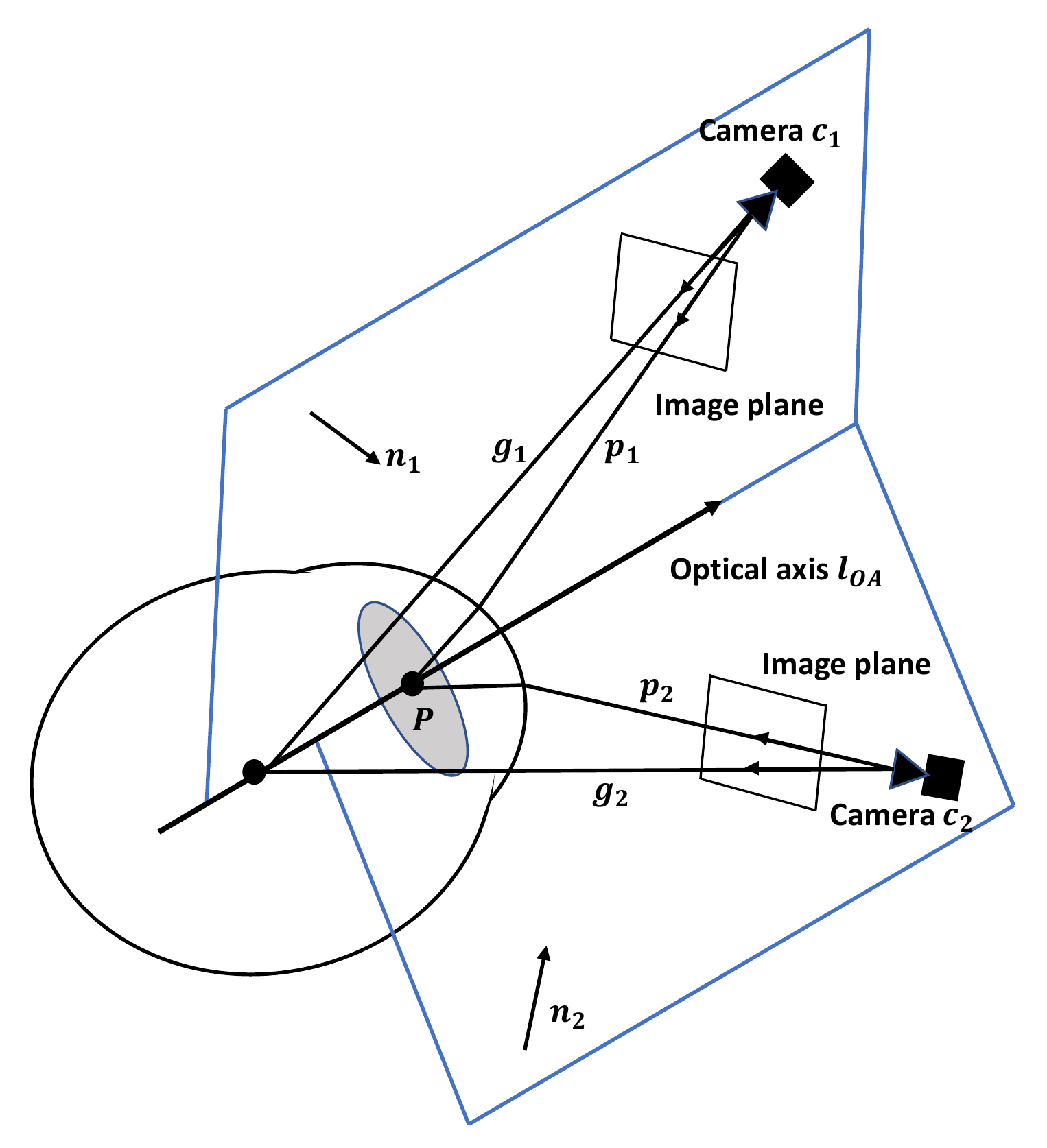} }

	\caption{Illustration of the near-eye-tracking.}
	\label{51020}
\end{figure}
According to the coordinates of the pupil centre obtained by processing the eye image, based on the camera imaging calibration theory\cite{zhang2000flexible}, the 3D coordinates of  the pupil center satisfy the following relationship
\begin{equation}
    \left [
    \begin{array}{c}
      u_p  \\
      v_p  \\
      1
     \end{array}\right] = \frac{1}{P_z}A 
     \left [
     \begin{array}{c}
     P_x  \\
     P_y  \\
     P_z
     \end{array}\right]= \frac{1}{P_z}
     \left [
     \begin{array}{ccc}
      f_x & 0 & c_x\\
      0 & f_y & c_y\\
      0 & 0 & 1
     \end{array}\right]
     \left [
     \begin{array}{c}
     P_x  \\
     P_y  \\
     P_z
     \end{array}\right],
\end{equation}
where $[u_p,v_p]$ is the gaze point estimated on the near-eye image, $[P_x,P_y,1]$ is the gaze point in the world coordinate system. $P_z$ and $A$ are the coordinate transformation matrix and the camera internal parameter matrix, respectively, which can be implemented by OpenCV \cite{bradski2000opencv}. To simplify the computation, the distance between the image plane and the camera is usually taken to be 1. Therefore, the 3D coordinates of the pupil are
\begin{equation}
	\left [
	\begin{array}{c}
	P_x  \\
	P_y  \\
	P_z
	\end{array}\right] =
		\left [
	\begin{array}{c}
	 P_x \\
	 P_y  \\
	 1
	 \end{array}\right]=
	 	\left [
	 \begin{array}{c}
	 \frac{u_p-c_x}{f_x} \\
	 \frac{u_p-c_y}{f_y}  \\
	 1
	 \end{array}\right].
\end{equation}
Similarly, the 3D coordinates of the centre of spot can also be found.  Thus, the direction vectors $p_1, p_2$ are calculated by the connection between the camera optical center and the pupil center. The direction vectors $g_1,g_2$ are derived from the lines between the camera center and the corneal spot center. Since the ocular optic axis, reflected corneal spot, and the center of eye camera are coplanar, the normal vectors of the two planes shown in Fig. \ref{51020} are calculated through

\begin{equation}
	\begin{cases}
    \bf{n_1}=\frac{  \bf{p_1} \times \bf{g_1}}{ \| \bf{p_1} \times \bf{g_1} \|} \\
    \bf{n_2}=\frac{  \bf{p_2} \times \bf{g_2}}{ \| \bf{p_2} \times \bf{g_2} \|}.
	\end{cases}
\end{equation}

The optical axis of the eye is the intersection of two planes, so the direction of the optical axis $l_{OA}$ can be obtained from the normal vectors of the two planes through equation $\frac{n_1\times n2}{ \|n_1\times n2\|}$,  after which the corresponding coordinates of the other positions of the optical axis can be calculated.

\begin{figure}[H]
	\centering
	\includegraphics[width=0.45\textwidth]{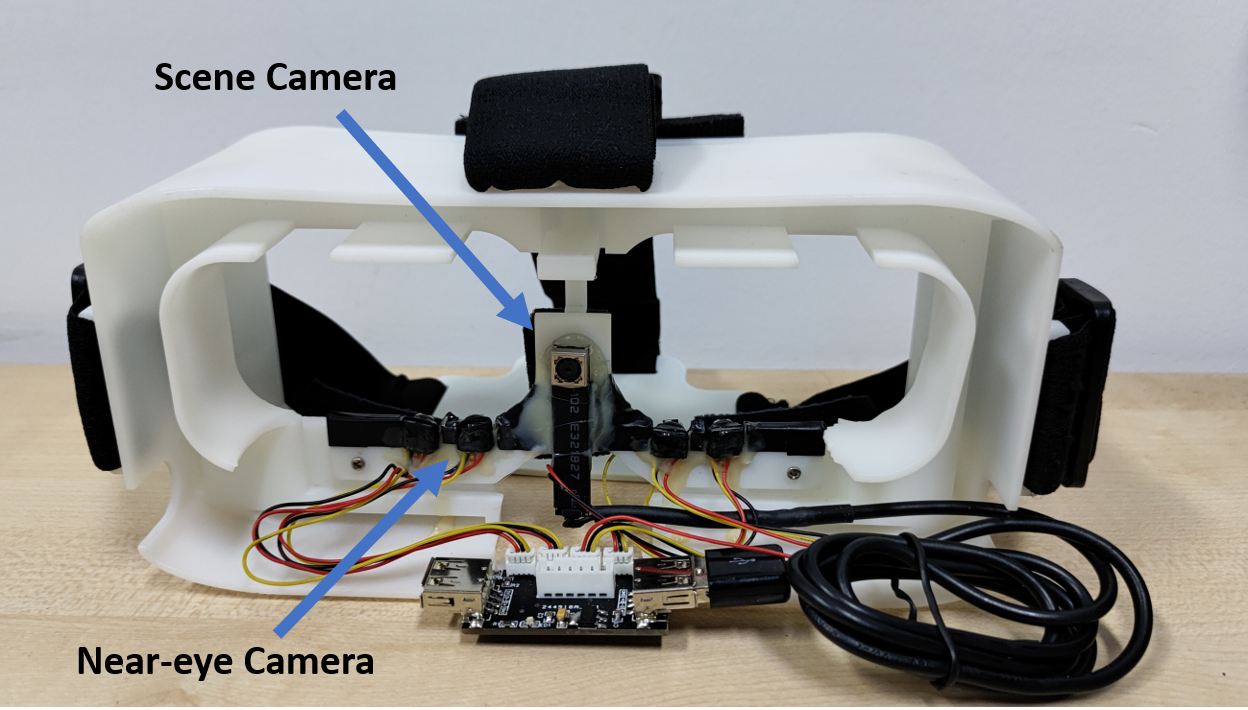} 
	\caption{The developed head-mounted eye-tracking device.}
    \label{dev}
\end{figure}

Our near-eye assistance system as a whole is shown in Fig. \ref{dev}, where  tiny eye cameras are installed on a head-mounted frame to capture high-resolution near-eye images. 
As shown in Fig. \ref{51020}, for each eye,  two near-eye cameras are used to capture the corneal reflected spot and calculate the corneal spherical center and 3D optical axis for the coordinates of the imaging.

At the same time, a scene image with the same view as the user is acquired by a scene camera mounted on the same frame.
The system builds on the spherical cornea model and applies multiple near-eye cameras to resolve the corneal center and optical axis of the eyes in 3D space, detects the pupil center, and fits the line of sight. The line of sight is solved by analyzing the eyeball and applying optical principles. The pupil's 3D coordinates are  obtained by detecting the centres  of pupil in the image from near-eye cameras. Since the optical axis of eyes passes through the center of pupil and the sphere of the cornea, the line of sight can be obtained by calculating the line between the two coordinates based on the geometric model of the eye \cite{nagamatsu2010gaze}.

\begin{figure}
	\centering
	\includegraphics[width=0.49\textwidth]{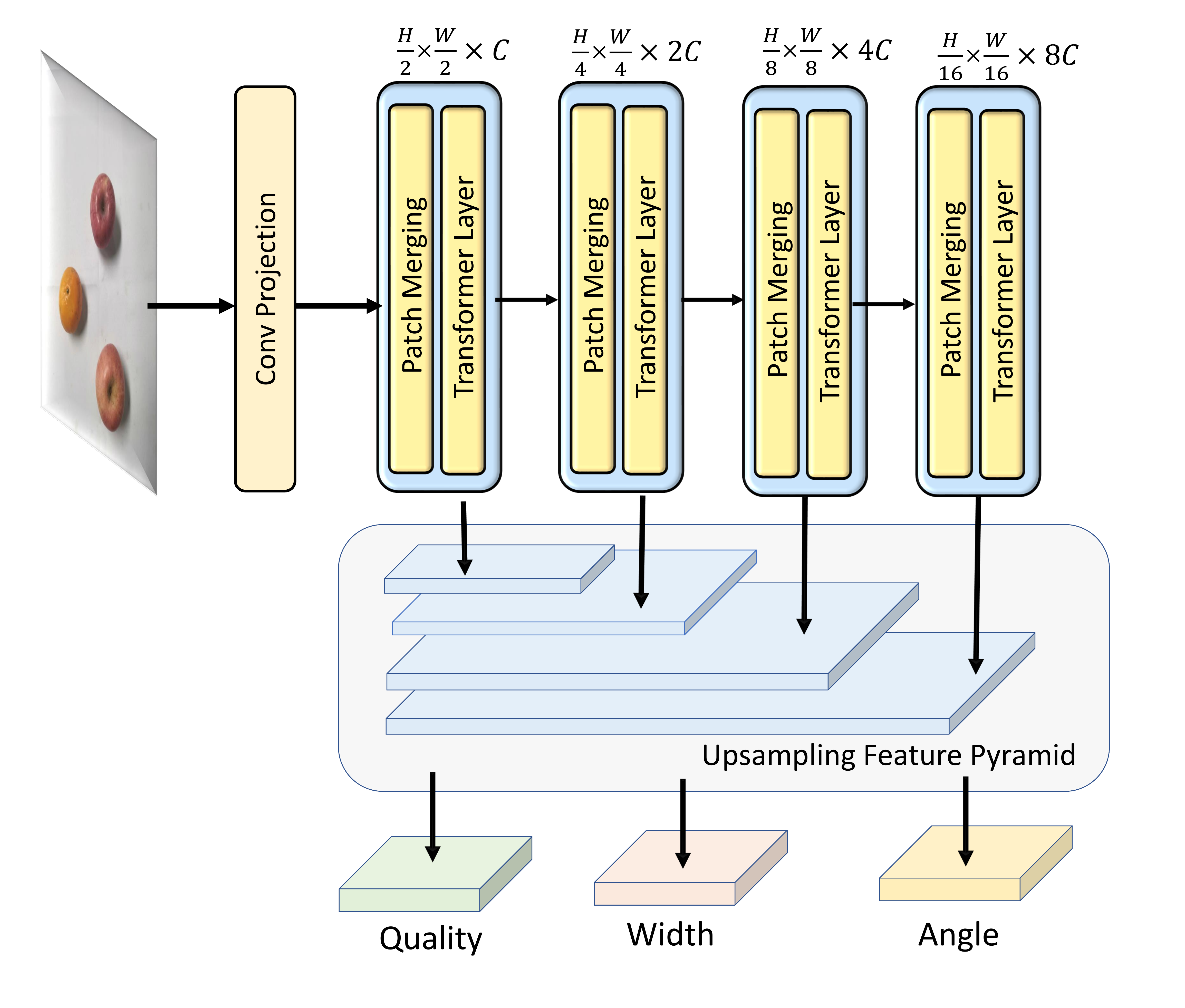} 
	\caption{Overview  of the transformer based grasp detection model.}
	\label{structure}
\end{figure}
\subsection{Grasp Detection}

Compared to object detection, the grasp detection is generally made of small rectangles and is more sensitive  to positions and rotation angles. To provide a more global understanding of the model, as opposed to convolutional kernels with fixed receptive fields, stacked transformer layers are adopted as backbone to gradually extract coarse-grained to fine-grained feature representations. As shown in Fig. \ref{structure}, the input image 
 $\mathcal{I} \in \mathbb{R}^{W \times H }$, where W and H are the width and height of the image $\mathcal{I}$. At first, the image is split into non-overlapping patches through a conv projection layer. Each patch in the image is treated as a word token. Similar to \cite{liu2021swin}, there are four successive blocks to extract semantic rich information, and each stage contains a patching merging layer and a swin transformer layer. Each block is composed of patch merging and swin transformer layer.
Patching merging is functionally similar to the pooling mechanism in CNN, which is designed to reduce the resolution of an image while increasing the number of channels of features. The dimension of features of each stage is shown in Fig. \ref{structure}. The foundation of swin transformer layer is the attention mechanism. Input features are linearly transformed to obtain query, key, and value.  Then the self-attention is computed as
\begin{equation}
\text{Attention}({Q},{K},{V})= \operatorname{SoftMax}(\frac{{Q} {K}^T}{\sqrt{d}}) {V},
\end{equation}
where $\sqrt{d}$ is the scale factor. Swin transformer layer performs self-attention within a local window, greatly reducing the  computational complexity. Meanwhile, the shifted window is applied to model the global relationships. The whole computation flow is as follows:
\begin{equation}
\begin{aligned}
\hat{\mathbf{x}}^{l} &=\operatorname{W-MSA}\left(\mathrm{LN}\left(\mathbf{x}^{l-1}\right)\right)+\mathbf{x}^{l-1}, \\
\mathbf{x}^{l} &=\operatorname{MLP}\left(\mathrm{LN}\left(\hat{\mathbf{x}}^{l}\right)\right)+\hat{\mathbf{x}}^{l}, \\
\hat{\mathbf{x}}^{l+1} &=\operatorname{SW-MSA}\left(\mathrm{LN}\left(\mathbf{x}^{l}\right)\right)+\mathbf{x}^{l}, \\
\mathbf{x}^{l+1} &=\operatorname{MLP}\left(\mathrm{LN}\left(\hat{\mathbf{x}}^{l+1}\right)\right)+\hat{\mathbf{x}}^{l+1}
\end{aligned}
\end{equation}
The feature $ \mathbf{x}^{l-1}$ from last layer enters the W-MSA module via the layerNorm layer, and there exists a residual connection between each module. After that,  it  goes through SW-MSA layer in a similar way. Here W-MSA represents the window multi-head attention layer and SW-MSA indicates shifted window multi-head attention layer. One motivation for using swin transformer as backbone network is that it maintains both global and local perception. Meanwhile, it reduces computational complexity compared to vanilla self-attention. 

A feature fusion pyramid is designed at the bottom of model in Fig. \ref{structure} to collect features from each stage in the backbone network. Features from different stages are aggregated through a feature pyramid for a multiscale fusion of contextual information. The feature fusion module uses concatenation to fuse these features that learn semantic and spatial contextual information. The network finally outputs the grasping  quality head, grasping width head, and grasping angle head via $1 \times 1$ convolutional kernels. Each output head is the same size as the original input. For grasping quality head, each parameter in the quality output takes a value between 0 and 1, indicating the probability of a successful grasp in the corresponding position in the image. The angle head consists of two parts: $cos2\theta$ and $sin2\theta$, and the resulting angle is calculated by $ \frac{1}{2} arctan\frac{cos2\theta}{sin2\theta}$.  Afterwards, all outputs of grasping quality head are searched for the point with the highest grasping quality as the grasping center, as well as its corresponding gripper rotation angle and width.

The loss function $\mathcal{L}$ is defined as $\mathcal{L}={w_1} \mathcal{L}_{pose}+ {w_2} \mathcal{L}_{angle}+{w_3}\mathcal{L}_{pose}$. For each component of the loss function, $\mathcal{L}_i$ is the mean square error between the corresponding value of the model and the ground truth. $w1, w2, $ and $w3$ are the relevant weight factors. For instance, the first term of $\mathcal{L}$ is defined as $\mathcal{L}_{pose}=\sum_{i=1}^N  \| \tilde{G}_i - {G}_i^*  \|^2$, where $\tilde{G}_i$ is the output of the grasp quality head and ${G}_i^*$ is the corresponding ground truth.
\begin{figure*}[]
	\centering
	
	\includegraphics[height=0.25\textwidth]{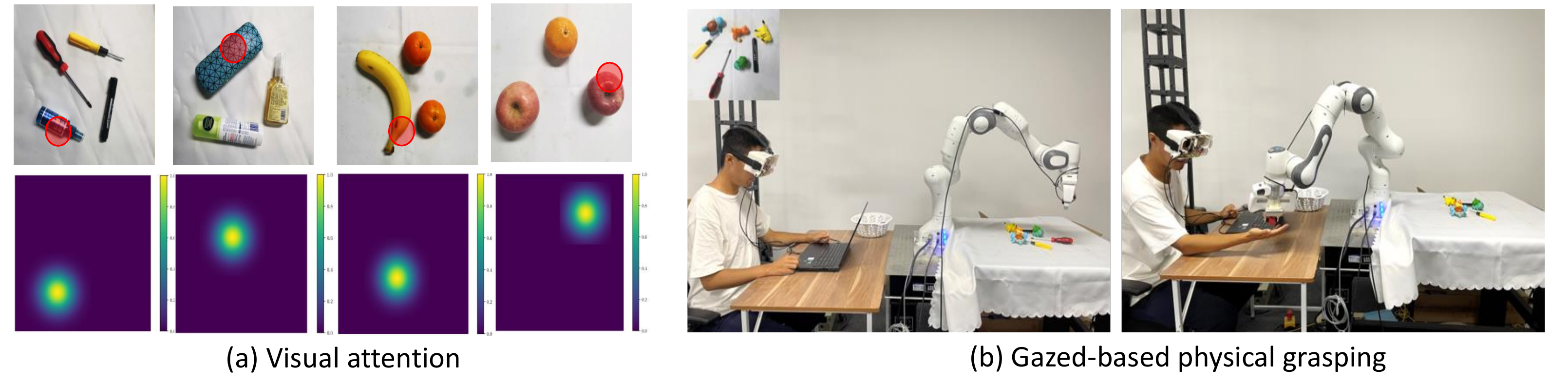} 
	\caption{ (a) Visualization of the scene images and the human attention heatmaps captured by our proposed method. From left to right, the human gaze points are the blue bottle, eyeglass box, banana, and apple, respectively. (b) Experiment for real-time gaze-based grasping.}
\end{figure*}
\begin{table*}[]
	\centering
	\setlength\tabcolsep{4pt}
	{
		\small
		\caption{ Experimental robotic grasping success rate for different objects. }
		
		\begin{tabular}{c|cc|c|cc| c|cc}
			
			\hline	\multicolumn{3}{c|}{Seen Objects}&	\multicolumn{3}{c|}{Familiar Objects}&	\multicolumn{3}{c}{Unseen Objects} \\ 
			\hline \text { Objects } & \text { Detected } & \text {  grasp (\%) }&
			\text { Objects } & \text { Detected } & \text {  grasp (\%) }&
			\text { Objects } & \text { Detected } & \text {  grasp (\%) } \\
			\hline \text { Mouse } & 15 / 15 & 13 / 15  &\text { Orange } & 14 / 15 & 12 / 15 & \text { Scissor } & 12 / 15 & 11 / 15\\
			
			\hline \text { Remote Control } & 15 / 15 & 13 / 15  &\text { Staples Box } & 15 / 15 & 12 / 15 & \text { Toothpaste Box } & 12 / 15 & 12 / 15\\

			\hline \text { Apple } & 14 / 15 & 12 / 15  &\text { Knife } & 12 / 15 & 11 / 15 & \text { Razor } & 13 / 15 & 11 / 15\\
			
			\hline \text { Pencil } & 14 / 15 & 13 / 15  &\text { Screwdriver } & 12 / 15 & 14 / 15 & \text { Toy } & 12 / 15 & 9 / 15\\
			\hline \text { Average } & 96 \% & 85 \%  &\text { Average } & 88 \% &81.6 \% &\text { Average } & 81.6 \% & 71.6 \%  \\
			\hline
			
		\end{tabular}
		\label{house_table}
	}
\end{table*}

\section{Experiment}

\subsection{Dataset and Implementation Details }
The Cornell \cite{lenz2015deep} grasping dataset  is used to evaluate the effectiveness of our grasp detection model. Each image in the dataset has been taken with a center crop of $224 \times 224$. The full grasp detection model is implemented in Pytorch, running on a single NVIDIA GTX 3090 GPU. The batch size is set to 32 and we use AdamW optimizer with a learning rate of $1e$-4.

\textbf{Evaluation Criteria.}
Following the standards in \cite{lenz2015deep},\cite{kumra2017robotic},\cite{wang2016robot}, the grasping rectangle metric is used  to evaluate the grasping results. A predicted grasp is treated as positive if it meets the following two criteria.

\romannumeral1 ) The discrepancy between  the rotation angle of the predicted grasp and the ground truth does not exceed $30^{\circ}$.

\romannumeral2) The Jaccard index, defined in \eqref{Jaccard index}, of the predicted grasp and the ground truth is greater than $0.25$,
\begin{equation}
\label{Jaccard index}
J\left(\mathcal{R}^{*}, \mathcal{R}\right)=\frac{\left|\mathcal{R}^{*} \cap \mathcal{R}\right|}{\left|\mathcal{R}^{*} \cup \mathcal{R}\right|},
\end{equation}
where $\mathcal{R}$ is the predicted grasping rectangle region and  $\mathcal{R}^{*}$ is the ground truth. $\mathcal{R}^{*} \cap \mathcal{R}$ is an intersection of these two areas and $\mathcal{R}^{*} \cup \mathcal{R}$ is the union of these two regions.
\subsection{Grasping Results}
Given the desired grasping coordinates, the robot inverse kinematics are utilized to realize the desired trajectory.
The detailed comparisons with other methods on the Cornell dataset are listed in Table \ref{ cornell results}. Although the performance gap among the state-of-the-art methods is moderate, our model achieves the best performance.  For instance, our transformer-based grasping model achieves an accuracy of 96.28 if only depth images are used as input, and 98.86 for the RGB-D as input. In particular, our model directly predicts the grasping quality, angle, and width of grasping rectangles, obviating the requirements for design anchors for different targets. 
\begin{table}[]	
	\begin{center}
		\setlength\tabcolsep{7pt}
		\caption{The accuracy on Cornell grasping dataset. } 
		\begin{tabular}{l|l|cc} 
			Authors & Algorithm & { Accuracy (\%) } &  \\
			
			\hline Jiang \cite{jiang2011efficient} & Fast Search & $60.5$ \\
			Lenz         \cite{lenz2015deep} & SAE, struct. reg. & $73.9$   \\
			Redmon       \cite{redmon2015real} & AlexNet, MultiGrasp & $88.0$  \\
			Wang \cite{wang2016robot} & Two-stage closed-loop & $85.3$  \\
			Asif \cite{asif2017rgb} & STEM-CaRFs & $88.2$   \\
			Kumra \cite{kumra2017robotic} & ResNet-50x2 & $89.2$   \\
			Morrison \cite{morrison2020learning} & GG-CNN & $73.0$  \\
			Guo \cite{guo2017hybrid} & ZF-net & $93.2$  \\
			Zhou \cite{zhou2018fully} & FCGN, ResNet-101 & $97.7$  \\
			Karaoguz \cite{karaoguz2019object} & GRPN & $88.7$  \\
			Asif \cite{asif2018graspnet} & GraspNet & $90.2$  \\
			\hline & GraspFormer-D & $96.28$   \\
			Our & GraspFormer-RGB & $97.72$  \\
			& GraspFormer-RGB-D & $\mathbf{98.86}$   \\
			\hline
		\end{tabular}
		\label{ cornell results}
	\end{center}
\end{table}

To test whether our model can be generalized to new scenes, objects are arranged in new positions with different orientations. We divide the objects into three categories, including objects that appear in the dataset, objects that are similar in the dataset, and objects that have never been seen before. In each category, there are at least four objects. Each type of object is grasped several times, and the number of successful grasps are recorded.   The detailed grasping results are shown in  Table \ref{house_table}. In Table \ref{house_table}, we can see that it performs well for tools seen in the dataset, and shows good generalization to similar objects. For unseen objects and complex scenes, our method also provides a decent improvement in grasp detection accuracy.

\subsection{System Design}

The system consists of two main components, the eye-tracking module and grasping module. The robot employed in our experiments is  a Franka Emika Panda robot. An RGB-D camera is fixed on the  robot's gripper.  The used camera is RealSense D435i.  The panda robot has a parallel finger and its active range is limited to $10$cm with a maximum loading capacity of no more than 3kg.

The eye-tracking hardware includes four eye cameras associated with infrared light source, one scene camera, multi-channel video acquisition circuits, and a head-mounted frame. The near-eye cameras are positioned below the human eyes to take high-resolution images of eyes, and the near-infrared light around the camera provides a corneal reflective spot used to accurately resolve the 3D eye features under the assumption of an aspheric corneal model.

\subsection{Limitations}
Since properties such as corneal refraction and corneal asphericity can introduce additional parameters which are hard to calibrate, it is challenging to directly solve for pupil centre coordinates by optical principles. To mitigate the challenge, a simplified and approximated eye model is used in our eye-tracking module. Specifically, the refraction of the cornea is ignored, whereas the refractive index of human cornea is approximately $1.376$. The curvature of the corneal surface is also overlooked, which is modelled as a sphere in our model. Note that in reality the curvature is not constant at all points on the surface of the eyes. Such  approximations can lead to  discrepancies in the gaze estimation. In the grasping subsystem, our model does not perform well when grasping transparent objects, since the RealSense camera does not yet provide a good depth for such objects. It is found in the experiment that the objects with complex surfaces or smooth materials are likely to slip out of the grippers during grasping.

\section{Conclusions and Discussions}

\begin{table}[h]
	\centering
	\setlength\tabcolsep{7pt}
	{
		\small
		\caption{Time Delay Analysis of gaze interaction-based grasping assistive system.  }
		\begin{tabular}{c|c|c}
			\hline \multicolumn{2}{c|}{\textbf{Setup}} & \textbf{ Time Delay} \\  \hline
			\multicolumn{2}{c|}{Image Acquisition} & 0.01 ms \\ 
			\multicolumn{2}{c|}{Image Preprocessing} & 0.2 ms \\ 
			\multicolumn{2}{c|}{Gaze Point Estimation} & 3 ms \\ \hline
			\multirow{3}{*}{grasp detection  } &  GraspFormer  Tiny   &  47 ms\\ 
			&   GraspFormer Small   &   56 ms \\
			&   GraspFormer Base   &   88 ms \\    \hline
			\multirow{3}{*}{Total Time   } &    GraspFormer Tiny   &  50.21ms\\ 
			&    GraspFormer Small   &   59.21ms \\
			&     GraspFormer Base   &   91.21ms \\    \hline

		\end{tabular}
		\label{time}
	}
\end{table}
\begin{figure}[h]
	\centering
	
	\includegraphics[width=0.44\textwidth]{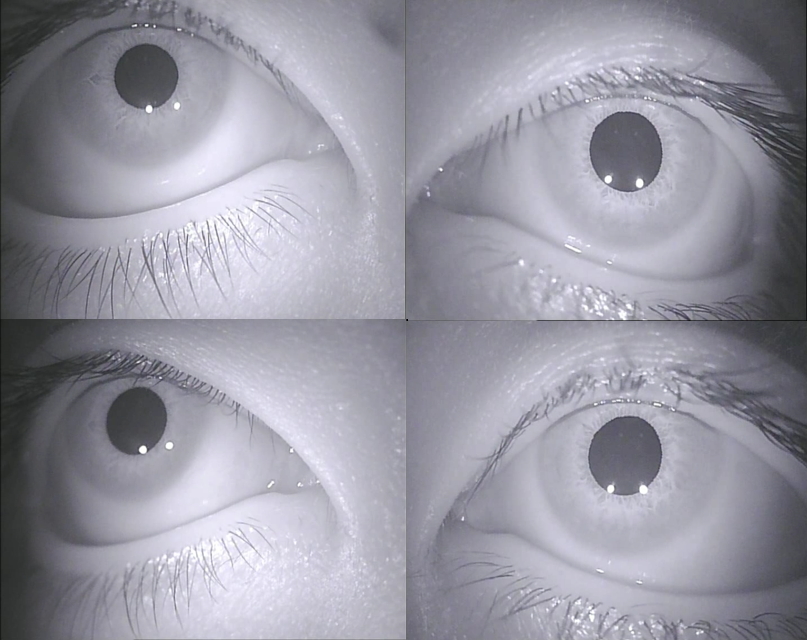}
	
	\caption{Images of the eye captured by near-eye cameras.
		\label{near-eye}}
\end{figure}
In this work, we provide a contactless human-robot interface that enables robotic manipulation using sight. The proposed framework leverages a head-mounted eye-tracking device to automatically locate the object that human pays attention to. Once the user's gaze point is obtained, a transformer-based grasp model takes advantages of global perception to identify the user's region of interest. The results show that the developed gaze-based robotic arm is capable of moving objects or grasping desired objects by near-eye-tracking. Ablation studies demonstrate that our eye-tracking can achieve reasonably decent tracking accuracy. The associated head-mounted device provides a low-cost design and meets real-time requirements.

Compared to using  other human-machine interfaces, our eye interaction is more flexible and user-friendly.
Eye-driven human-robot interaction serves as a novel framework and shows great potential in various applications. Future research will consider extending the current work to more challenging environments (e.g., outdoor environments) for more complex tasks (e.g., pouring water, serving food).


\ifCLASSOPTIONcaptionsoff
  \newpage
\fi



%


\bibliographystyle{IEEEtran}
\bibliography{eye.bib}
%





\end{document}